\pdfoutput=1
\documentclass{article}
\usepackage[pdftex]{graphicx}
\usepackage{amsthm} 
\usepackage{amsmath}
\usepackage{amssymb}
\usepackage{amsfonts}
\usepackage{amsbsy}
\usepackage{array}
\usepackage{subfigure}
\usepackage{float}
\usepackage{amstext}
\usepackage{epsfig}
\usepackage{color}
\usepackage{multirow}
\usepackage{slashbox}
\usepackage{enumerate}
\usepackage{fullpage}
\usepackage{authblk}

\DeclareMathOperator*{\argmin}{argmin}

 \newtheorem{definition}{Definition}
\newtheorem{fact}{Fact}
 \newtheorem{lemma}{Lemma}
 \newtheorem{theorem}{Theorem}
 \newtheorem{corollary}{Corollary}

\newcommand{\etal}{{et al. }}

\newcommand{\hx}{\hat{x}}

\newcommand{\hy}{\hat{y}}

\newcommand{\hb}{\hat{b}}
\newcommand{\hf}{\hat{f}}
\newcommand{\hg}{\hat{g}}

\newcommand{\hv}{\hat{v}}

\newcommand{\bb}{\bar{b}}

\newcommand{\bv}{\bar{v}}

\newcommand{\cC}{{\mathcal C}}
\newcommand{\cE}{{\mathcal E}}
\newcommand{\cG}{{\mathcal G}}

\newcommand{\cL}{{\mathcal L}}

\newcommand{\cO}{{\mathcal O}}

\newcommand{\cV}{{\mathcal V}}

\newcommand{\realset}{{\mathbb R}}
\newcommand{\intset}{{\mathbb Z}}

\newcommand{\beq}{\begin{equation}}
\newcommand{\eeq}{\end{equation}}

\newcommand{\beqa}{\begin{eqnarray}}
\newcommand{\eeqa}{\end{eqnarray}}

\newcommand{\balgn}{\begin{align}}
\newcommand{\ealgn}{\end{align}}

\newcommand{\bmtln}{\begin{multline}}
\newcommand{\emtln}{\end{multline}}

\newcommand{\bitm}{\begin{itemize}}
\newcommand{\eitm}{\end{itemize}}

\newcommand{\benum}{\begin{enumerate}}
\newcommand{\eenum}{\end{enumerate}}

\newcommand{\bary}{\begin{array}}
\newcommand{\eary}{\end{array}}

\newcommand{\bdefn}{\begin{definition}}
\newcommand{\edefn}{\end{definition}}

\newcommand{\bfct}{\begin{fact}}
\newcommand{\efct}{\end{fact}}

\newcommand{\blem}{\begin{lemma}}
\newcommand{\elem}{\end{lemma}}

\newcommand{\bthm}{\begin{theorem}}
\newcommand{\ethm}{\end{theorem}}

\newcommand{\bcor}{\begin{corollary}}
\newcommand{\ecor}{\end{corollary}}

\newcommand{\bassum}{\begin{assumption}}
\newcommand{\eassum}{\end{assumption}}

\newcommand{\bprf}{\begin{proof}}
\newcommand{\eprf}{\end{proof}}

\usepackage[ruled,lined]{algorithm2e}

\newcommand{\solutionset}{characteristic set}

\newcommand{\SolutionSet}{Characteristic Set}

\newcommand{\cc}{\textcolor{black}}

\newcommand{\algname}{{\sc DualSearch}}
\newcommand{\algoneA}{{\sc DualMax}}
\newcommand{\algoneB}{{\sc AdaptSearch}}

\newcommand{\algnamec}{{\sc DualSearch }}
\newcommand{\algoneAc}{{\sc DualMax }}
\newcommand{\algoneBc}{{\sc AdaptSearch }}

\title{Multi-dimensional Parametric Mincuts for Constrained MAP Inference}

\author[1]{Yongsub Lim\thanks{yongsub@kaist.ac.kr}}
\author[1]{Kyomin Jung\thanks{kyomin@kaist.edu}}
\author[2]{Pushmeet Kohli\thanks{pkohli@microsoft.com}}
\affil[1]{Korea Advanced Institute of Science and Technology}
\affil[2]{Microsoft Research Cambridge}

\date{}

\begin{document}

\maketitle

\begin{abstract}
In this paper, we propose novel algorithms for inferring the Maximum a Posteriori (MAP) solution of
discrete pairwise random field models under multiple constraints. We show how this constrained
discrete optimization problem can be formulated as a multi-dimensional parametric mincut problem via
its Lagrangian dual, and prove that our algorithm isolates all constraint instances for which the
problem can be solved exactly. These multiple solutions enable us to even deal with `soft
constraints' (higher order penalty functions). \cc{Moreover, we propose two practical variants of
our algorithm to solve problems with hard constraints.} We also show how our
method can be applied to solve various constrained discrete optimization problems such as submodular
minimization and shortest path computation. Experimental evaluation using the foreground-background
image segmentation problem with statistic constraints reveals that our method is faster and its
results are closer to the ground truth labellings compared with the popular continuous relaxation
based methods.
\end{abstract}

\section{Introduction}

Markov Random Fields (MRF) is an undirected graphical model, which has been extensively
studied and used in various fields, including statistical physics \cite{KindermannS80},
and computer vision \cite{Li94}. It represents interdependency of discrete random
variables as a graph over which a probabilistic space is defined. Computing the solution
which has the maximum probability under the random field, or Maximum a Posteriori (MAP)
inference is NP-hard in general. However, a number of subclasses of MRFs have been
isolated for which the problem can be solved in polynomial time \cite{borosdam02}.
Further, a number of heuristics or approximation algorithms based on belief
propagation \cite{weissuai07}, tree reweighted message passing \cite{wainwrightit05},
and graph-cut \cite{boykovpami01} have also been proposed for the problem.
Such algorithms are widely used for various problems in machine learning and computer
vision \cite{YanoverMW06,LadickyRKT10}. Since MAP inference in an MRF is equivalent to
minimizing the corresponding energy function\footnote{Energy of a labelling is the negative
logarithm of its posterior probability.}, in what follows, we will explain \cc{these problems}
in terms of energy minimization.

In many real world problems, the values of certain statistics of the desired solution may be
available as prior knowledge. For instance, \cc{in the case of foreground-background
image segmentation,} we may know the approximate shape and/or size of the object being
segmented, and thus might want to find the most probable segmentation that has a particular
area (number of foreground pixels) and boundary length (number of discontinuities).
Another example is \cc{community detection} in a network \cite{Hastings06}
where we may know the number of nodes belonging to each community.
\cc{Such scenarios result in constraints in the solution space, and MAP inference becomes
a constrained energy minimization problem, which is generally NP-hard even if the unconstrained version
is polynomial time solvable.}

Energy minimization under the above-mentioned statistics constraints results in a challenging
optimization problem. However, recent work in computer vision has shown that this problem can be
handled efficiently using the parametric mincuts \cite{KolmogorovBR07} which allow simultaneous
computation of exact solutions for some constraint instances. Although the parametric mincuts
provide a general framework to deal with constrained energy minimization, they can only handle one
linear equality constraint.

For minimizing energy functions under multiple constraints, a number of continuous relaxation based
methods have been proposed in the literature. For instance, linear relaxation approaches were
adopted to handle bounding-box and connectivity constraints defined on the
labelling \cite{lempitskyiccv09,nowozincvpr09}.
Further, Klodt and Cremers \cite{KlodtC11} proposed a convex relaxation framework to deal with moment constraints. 
Continuous relaxation based methods have also been used for constrained discrete
optimization, and can handle multiple inequality constraints. All the above-mentioned methods 
suffer from following basic limitations: they only handle linear constraints, and the solution 
involves rounding of the solution of the relaxed problem which may introduce large errors.

\subsection{Our contribution}

In this paper we show how the constrained discrete optimization problem associated with constrained
MAP inference can be formulated as a multi-dimensional parametric mincut problem via its Lagrangian
dual, and propose an algorithm that isolates all constraint instances for which the
problem can be solved exactly. This leads to \cc{\em densely} many minimizers, each of which is,
optimal under distinct constraint instance. These minimizers can be used to compute good approximate
solutions of problems with soft constraints (enforced with a higher order term in the energy).

Our algorithm works by exploiting the Lagrangian dual of the minimization problem, and requires an
oracle which can compute values of the Lagrangian dual efficiently. A graph-cut
algorithm \cite{boykovpami01} is a popular example of such an oracle. In fact, our algorithm
generalizes the \cc{(one-dimensional)} parametric mincuts \cite{gallo89,KolmogorovBR07} to multiple-dimensions.
In contrast to the parametric mincuts \cite{gallo89}, our algorithm can deal with multiple constraints
simultaneously, including some non-linear constraints (as we show in the paper).
\cc{This extension allows our algorithm to be used as a technique for multi-dimensional sampling
e.g.~to obtain different segmentation results for image segmentation as done in \cite{CarreiraS12}.}

We propose two variants of our algorithm to efficiently deal with the problem of performing 
MAP Inference under hard constraints. The first variant computes the maximum of the dual and outputs 
its corresponding primal solution as an approximation of the constrained minimization. \cc{The primal is 
computed using selective oracle calls, leading to fast computation time.} The other variant
combines the first variant with our multi-dimensional parametric mincuts algorithm to deal with problems with soft-constraints, \cc{which allows to find a solution closer to a desired one via additional search.}

Our method is quite general and can be applied to any constrained discrete optimization
problems whose Lagrangian dual value is efficiently computable. Examples include
submodular function minimization with constraints such as the balanced minimum cut
problem, and constrained shortest path problems. Further, in contrast to traditional 
continuous relaxation based methods, our technique can easily handle complicated soft constraints.

In Section \ref{sec:app}, we demonstrate that our algorithms compute solutions very close
to the ground truth compared with these continuous relaxation based methods on the foreground-background
image segmentation problem.




\subsection{Related work}

A number of methods have been proposed to obtain better labelling solutions by inferring
the MAP solution from a restricted domain of solutions which satisfy some constraints.
Among them, solutions to image labelling problems which have a particular distribution of
labels \cite{woodfordiccv09} or satisfy a topological property like
connectivity \cite{VicenteKR08} have been widely studied.

More specifically, for the problem of foreground-background  image segmentation, most probable
segmentations under the label count constraint have been shown to be closer to the ground
truth \cite{KolmogorovBR07,LimJK10}.
Another example is the silhouette constraint which has been used for the problem of 3D
reconstruction \cite{koleveccv08,sinhaiccv05}. This constraint ensured that a ray emanating from
any silhouette pixel must pass through at least one voxel which belongs to the `object'.

Recently, dual decomposition has been proposed for constrained MAP inference
\cite{GuptaDS07,WoodfordRK09}. Gupta \etal \cite{GuptaDS07} dealt with
cardinality-based clique potentials and developed both exact and approximate
algorithms. Also Woodford \etal \cite{WoodfordRK09} studied a problem involving
marginal statistics such as the area constraint especially with convex
penalties, and showed that the proposed method improves quality of solutions for
various computer vision problems.

MAP inference under constraints are also applied to combinatorial
optimization such as the balanced metric labelling. For this problem, Naor and Schwartz \cite{NaorS06}
obtained an $O(\tfrac{\ln{n}}{\epsilon})$-approximate algorithm where each label is
assigned to at most $\min\left[ \tfrac{O(\ln{k})}{1-\epsilon}, \ell+1
\right](1+\epsilon)\ell$ variables/nodes.




\section{Setup and preliminaries} \label{sec:setup}

\subsection{Energy minimization}

Markov Random Fields (MRF) defined on a graph $G=(V,E)$ is a probability distribution
where every vertex $u\in V$ has a corresponding random variable $x_u$ taking a value from the
finite label set $\cL$. The probability distribution is defined as $\Pr(x)
\propto \exp(-f(x))$ where $x=(x_u)$, and the corresponding energy function $f$ is in the
following form:
\beq \label{eq:E}
	f(x) = \sum_{c\in \cC_G} \phi_c(x_c),
\eeq
where $\cC_G$ is the set of cliques in $G$ and $\phi_c$ is a {\em potential} defined over the
clique $c$. The MAP problem is to find an assignment $x^* \in \cL^{|V|}$ which has the
maximum probability, and is equivalent to minimizing the corresponding energy function
$f$. In general it is NP-hard to minimize $f$, but it is known that if $f$ is submodular,
it can be minimized in polynomial time. Especially, if $f$ is a pairwise submodular energy
function defined on binary variables, which considers only cliques of size up to $2$, i.e.
\beq \label{eq:pairwise_energy_function}
f(x) = \sum_{u\in V} \phi_u(x_u) + \sum_{(u,v)\in E} \phi_{uv}(x_u,x_v),
\eeq
it can be efficiently minimized by solving a equivalent
st-mincut problem~\cite{KolmogorovR09}. 
\cc{Such $f$}
is widely used in machine learning and computer vision~\cite{Ishikawa11,LimJK10}.

\subsection{Energy minimization with constraints}

Energy minimization with constraints is to compute the solution $x^*$ minimizing an energy
function among $x$'s satisfying given constraints. A typical example of constraints is the
label count constraint $\sum_i x_i=b$ where $x_i\in\{0,1\}$.

In this paper, we consider the following energy minimization with multiple constraints.
\beq \label{eq:our_problem}
	\min_{x\in\{0,1\}^n} \left\{ f(x) : h_i(x) = b_i, ~ 1\leq i\leq m \right\},	
\eeq
where $x\in \{0,1\}^n$, $m$ is a constant, and for $1\leq i\leq m$,
$h_i:\{0,1\}^n\rightarrow \realset$ and $b_i\in \realset$.
In \eqref{eq:our_problem}, each constraint $h_i(x) = b_i$ encodes \cc{distinct} prior knowledge on
a desired solution.
For convenience, we denote $\left(h_1(x),\ldots,h_m(x)\right)$ by $H(x)$.

Let us consider the following Lagrangian dual $g:\realset^m\rightarrow \realset$ of $f$, 
which is widely used for discrete optimization \cite{Komodakis07,StrandmarkK10}.
\beq\label{eq:dual}
	g(\lambda) = \min_{x\in \{0,1\}^n} L(x,\lambda),
\eeq
where
\beq\label{eq:lag}
	L(x,\lambda) = f(x) + \lambda^T (H(x)-b).
\eeq
Note that $g$ is defined over a continuous space while $f$ is defined over a discrete
space.
As in the continuous minimization, maximizing $g$ over $\lambda\in\realset^m$ provides a
lower bound for \eqref{eq:our_problem}. Now we define the {\em characteristic set}, which
is the collection of {\em minimizers} of \eqref{eq:lag} over all $\lambda\in\realset^m$.
\bdefn[\SolutionSet]
	The \SolutionSet~is defined by
	\beq
		\chi_g = \bigcup_{\lambda\in \realset^m} \argmin_{x\in\{0,1\}^n} L(x,\lambda).
	\eeq
\edefn
\blem
	Let $x^* \in \chi_g$ and $b^*=H(x^*)$. Then
	$f(x^*) = \min_{x\in\{0,1\}^n} \left\{ f(x) : H(x)\right.$ \\  $\left.=b^* \right\}$ \cite{Guignard03}.
\elem
\bprf
	Suppose that $\hx$ satisfies that $H(\hx)=b^*$. It implies
	$\lambda^T(H(\hx)-b^*) = \lambda^T(H(x^*)-b^*)$ for any $\lambda\in \realset^m$.
	Since $x^*\in \chi_g$, $L(x^*,\lambda)\leq L(\hx,\lambda)$
	for some $\lambda\in \realset^m$. Thus, from \eqref{eq:lag}, $f(x^*)\leq f(\hx)$.
\eprf

In this paper, we develop a novel algorithm to compute the
\solutionset~$\chi_g$.
We will show that if the dual $g(\lambda)$ is efficiently
computable for any fixed $\lambda\in\realset^m$, for example, when $L(x,\lambda)$ is submodular on
$x$, our algorithm computes $\chi_g$ by evaluating $g(\lambda)$ for $poly(|\chi_g|)$
number of $\lambda\in\realset^m$. One implication of $\chi_g$ is
\beq
	g(\lambda) = \min_{x\in \{0,1\}^n} L(x, \lambda) = \min_{x\in \chi_g} L(x, \lambda),
\eeq
meaning that $\min_{x\in \{0,1\}^n} L(x,\lambda)$ indeed depends on a much smaller set
$\chi_g$. Note that $\chi_g$ does not depend on the constraint instance $b$, thus, in the
remaining of the paper, we regard $b = {\bf 0}$ unless there is explicit specification. In
Section \ref{sec:app}, we will show that $|\chi_g|$ is polynomially bounded in $n$ for
many constraints corresponding to useful statistics of the solution. Through experiments,
we will show that $|\chi_g|$ is densely many among all possible constraint instances by
an example of image segmentation.

Note that if we can compute minimizers of \eqref{eq:our_problem} for densely many
constraint instances $b$, we can obtain a good approximate solution
for the following soft-constrained problem with any \cc{global} penalty function $\rho$.
\beq\label{eq:soft-constrained_problem}
	\min_{x\in\{0,1\}^n} \left\{ f(x) + \rho(H(x) - \hb) \right\}.
\eeq
In \eqref{eq:soft-constrained_problem}, $\hb$ encodes our prior knowledge on a solution,
and examples  of $\rho$ include $\Vert \cdot \Vert_{\ell_p}$ and sigmoid functions.
This soft-constrained optimization has been widely used in terms of lasso regularization and
ridge regression, and also in computer vision~\cite{LadickyRKT10,ToyodaH08}.


\subsection{Generalization}

Although we describe our method for problems involving
pseudo-Boolean\footnote{\cc{Real-valued} functions defined over boolean vectors $\{0,1\}^n$.} objective
functions, there is a class of multi-label functions to which our method can be
applied. For instance, the results of \cite{RamKAT08} show transformation of any
multi-label submodular functions of order up to $3$ to a pairwise submodular one, meaning
that it can be solved by the graph-cut algorithm. This enables us to handle the following
type of constraints, which is analogue of linear constraints in binary cases:
for each $j\in\cL$,
\beq
	h_j(x) = \sum_{i\in V} a_{ij}\delta_{x_i;j} = b_j,
\eeq
where $\delta_{x_i,j}$ is Kronecker delta function.

Our method is also applicable to any constrained combinatorial optimization problems whose
$g(\lambda)$ is efficiently computable. We will discuss it more in detail in
Section~\ref{ssec:combinatorial_opt}.


\section{Computing the \SolutionSet} \label{sec:alg}
\subsection{Algorithm description}

In this section, we describe our algorithm that computes the
\solutionset~$\chi_g$. \cc{We assume that for a given set $S=\prod_{i=1}^m
[N_i,M_i]$ where $N_i,M_i\in\realset$ for all $i$, there is an oracle to compute
the Lagrangian dual $g$ efficiently for any $\lambda\in S$.} For simplicity of
explanation, we assume $S=[-M,M]^m$ for some $M>0$. We denote the oracle call by
\beq
	\cO(\lambda) = \argmin_{x\in \{0,1\}^n} L(x,\lambda).
\eeq
Essentially, our algorithm \cc{iteratively} decides the $\lambda$'s in $S$ for which
the oracle will be called. Later we prove that the number of oracle calls in our algorithm
to compute $\chi_g$ is polynomial in $|\chi_g|$.

We first define the following, which has a central role in our algorithm.
\bdefn[Induced dual of $g$ on $X$]
	Let $g:\realset^m\rightarrow \realset$ be the Lagrangian dual of $f$, and
	$X\subseteq\{0,1\}^n$.
	The induced dual $g_X$ of $g$ is defined by
	\beq
		g_X(\lambda) = \min_{x\in X} L(x,\lambda).
	\eeq
\edefn
From the definition of $\chi_g$, note that $g=g_{ \{0,1\}^n } = g_{\chi_g}$.
For each $x\in \{0,1\}^n$, we define a hyperplane $P_x$ by
\beq \label{eq:hyperplane}
	P_x = \{(\lambda,z)\in \realset^{m+1} : \lambda\in \realset^m, z=L(x,\lambda)\}.
\eeq
For $(\lambda,z)\in P_x$, we use the notation so that $P_x(\lambda)=z$.
For convenience, we will denote any $v\in \realset^{m+1}$ by $(\lambda_v, z_v)$,
where $\lambda_v\in \realset^m$ is the first $m$ coordinates of $v$ and $z_v\in
\realset$ is the $(m+1)$-th coordinate of $v$. Since $\{0,1\}^n$ is finite and
each $x \in \{0,1\}^n$ corresponds to a hyperplane in $(m+1)$-dimension, $g$
consists of the boundary of the upper polytope of \eqref{eq:dual}. Then
$\chi_g$ corresponds to the collection of $m$-dimensional {\em facets} of this
polytope.

To compute $\chi_g$, we will recursively update a structure called the {\em
skeleton} of $g_X$ defined below.
Intuitively, the skeleton of $g_X$ is the collection of {\em vertices} and {\em edges}
of the polytope corresponding to $g_X$.

\bdefn[Proper convex combination]
	Given $x,x_1,\ldots,x_k\in \realset^{\ell}$,
	$x$ is  a proper convex combination of
	$\{x_i : 1\leq i\leq k\}$ if $x = \sum_{i=1}^k \alpha_i x_i$ for some
	$\alpha\in (0,1)^k$ with $\sum_{i=1}^k \alpha_i = 1$.
\edefn

\bdefn[Skeleton of $g_X$ over $S$]
	For a given induced dual $g_X:\realset^m\rightarrow \realset$, let
	$\Gamma_X(S) = \{q\in \realset^{m+1} : \lambda_q \in S, ~z_q \leq
	g_X(\lambda_q)\}$, and for $u,v\in\Gamma_X(S)$, $e(u,v)\subseteq \Gamma_X(S)$
	is the line segment connecting $u$ and $v$. The skeleton of $g_X$ is
	$\cG_{g_X} = (\cV_{g_X},\cE_{g_X})$ satisfying the followings.
	\benum[\quad $\bullet$]
		\item \cc{$\cV_{g_X}=\{v\in\Gamma_X(S)~:~\text{if $v$ is a proper convex combination}$
		of $U\subseteq\Gamma_X(S)$, then $U=\{v\} \}$.}
		\item \cc{$\cE_{g_X} = \{e(u,v) : u,v\in\cV_{g_X},\text{ and if}~y\in e(u,v)$ is a
		proper convex combination of $W\subseteq \Gamma_X(S)$, then $W\subseteq e(u,v) \}$
		$~\cup~$ $\{e(u,v) : u\in\cV_{g_X}, ~ \lambda_u\in \{-M,M\}^m,~ v = (\lambda_u, -\infty) \}$.}	
	\eenum	
\edefn

Our algorithm runs by updating $X\subseteq \chi_g$ and $\cG_{g_X}$ iteratively.
If a new minimizer $x\in\{0,1\}^n$ is computed by the oracle call, it is
inserted to $X$ and the algorithm computes $\cG_{g_X}=(\cV_{g_X},\cE_{g_X})$.
\cc{Then, the algorithm determines new $\lambda$'s for which the oracle will be
called from the new vertices added to $\cV_{g_X}$.}
We prove in Theorem~\ref{thm:x_correct} that at the end of the algorithm, $X =
\chi_g$.

Initially, the algorithm begins with $X=\{x_0\}$ where $x_0$ is the output of
the oracle call for \cc{any arbitrary} $\lambda_0 \in \{-M,M\}^m$.
\cc{The inittial skeleton $\cG=(\cV,\cE)$ is given by $\cV=\{v_1,\ldots,v_{2^m}\}
\subset \realset^{m+1}$ where $\{\lambda_{v_i} : 1\leq i\leq 2^m\} =
\{-M,M\}^m$ and $z_{v_i} = P_{x_0}(\lambda_{v_i})$ for $1\leq i\leq 2^m$; and
$\cE = \cE_{g_X}$.}
Note that $\cG = \cG_{g_X}$, i.e.~the skeleton of $g_X$. This initialization is
denoted by $InitSkeleton()$ and it returns $X$ and $\cG$.

In each iteration with the skeleton $\cG_{g_X}=(\cV_{g_X},\cE_{g_X})$, the
algorithm chooses any vertex $v\in \cV_{g_X}$, and checks whether $z_v =
g_X(\lambda_v)$ using the oracle call for $\lambda_v$. If $z_v =
g_X(\lambda_v)$, we confirm that $z_v=g(\lambda_v)$ and $v\in \cV_g$. If not,
$x_v \notin X$ computed from the oracle satisfies $P_{x_v}(\lambda_v)
= g(\lambda_v)<z_v$. Then, the algorithm computes a new skeleton
$\cG_{g_{X\cup\{x_v\}}}$ as explained below.

Let $X'=X\cup\{x_v\}$. To compute $\cG_{g_{X'}}$, geometrically we cut $\cG_{g_X}$ by
$P_{x_v}$. This can be done by finding the set $\cV^-$ of skeleton vertices of
$g_X$ strictly above $P_{x_v}$, and finding the set $\cV^+$ of all intersection
points between $P_{x_v}$ and $\cE_{g_X}$. Then, $\cV^-$ is removed from
$\cV_{g_{X}}$, and $\cV^+$ is added to $\cV_{g_{X}}$.
Lastly, the set of edges of the convex hull of $\cV^+$, which is denoted by
$ConvEdge(\cV^+)$, is added to $\cE_{g_X}$\footnote{ For a given $\cV^+$,
$ConvEdge(\cV^+)$ can be computed, for example, by \cite{BarberDH96}. In
general, for given $(m+1)$-dimensional points, a convex hull algorithm outputs a
set of $m$ dimensional facets of the convex hull. Then, we can obtain the edges
of the convex hull by recursively applying the algorithm to every computed
facets.}. Then, \cc{the updated $\cG_{g_X}$ is $\cG_{g_{X'}}$.} Due to the
concavity of $g_X$, we can compute all the above sets by the depth or breadth first
search starting from $v$. Algorithm \ref{alg:alg_all} describes the whole
procedure.

\IncMargin{1em}
\begin{algorithm}[t]
\DontPrintSemicolon
\LinesNumbered

\BlankLine

\KwIn{Oracle $\cO$} \KwOut{$X$}


\BlankLine

$(X, \cG) \leftarrow InitSkeleton()$ \;
Give $\cV$ an arbitrary order \;

\ForEach{$v \in \cV$ in the order}{
	$x_v = \cO(\lambda_v)$ \;
	\If{ $P_{x_v}(\lambda_v) < z_v$ }{
		$X = X \cup \{x_v\}$ \;
		Append $\cV^+ = \{u \in P_{x_v}\cap e : e\in \cE, e \not\subseteq P_{x_v}
		\}$ to $\cV$ in arbitrary order \label{line:add_v}  \;
		Remove $\cV^- = \{u\in \cV : z_u > P_{x_v}(\lambda_u)\}$ from $\cV$\;
		$\cE^- = \{e(u_1,u_2)\in \cE : u_1\in\cV^- \text{ or } u_2 \in \cV^- \}$\;
		$\cE^+ = \{e(u_1,u_3) : \exists ~ e(u_1,u_2)\in \cE^-, ~ u_3 = e(u_1,u_2)\cap P_{x_v}  \}$\;
		$\cE = \cE \cup ConvEdge(\cV^+)\cup \cE^+ - \cE^-$\;
	}
}
\caption{\algname} \label{alg:alg_all}
\end{algorithm}\DecMargin{1em}



\begin{figure}
\centering
\subfigure[]{\includegraphics[width=0.45\columnwidth]{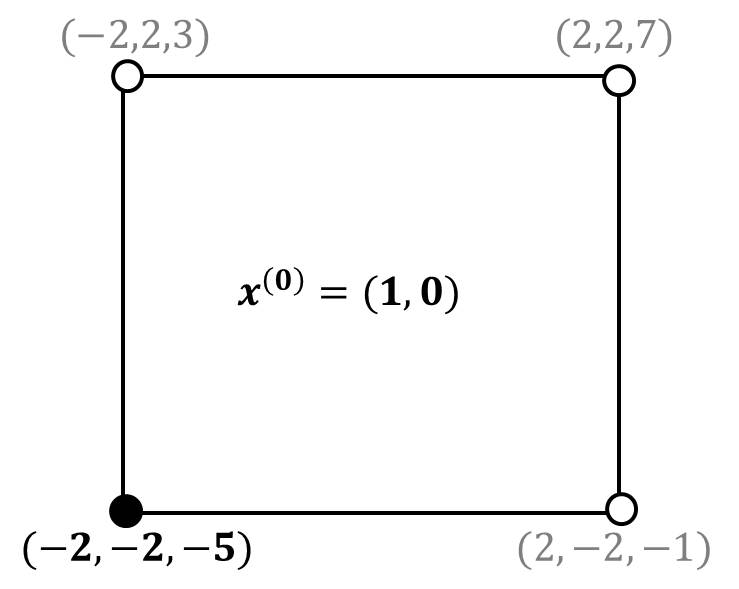}\label{fig:exe1}}
\subfigure[]{\includegraphics[width=0.45\columnwidth]{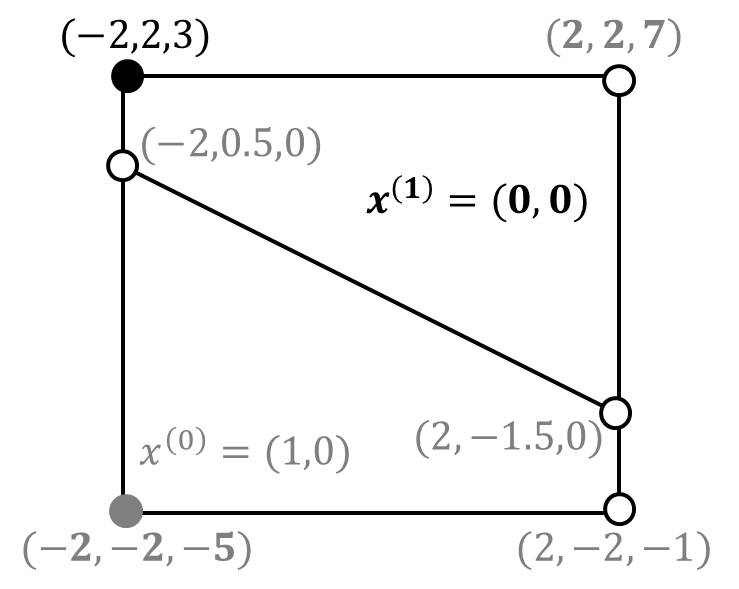}\label{fig:exe2}}
\\
\subfigure[]{\includegraphics[width=0.45\columnwidth]{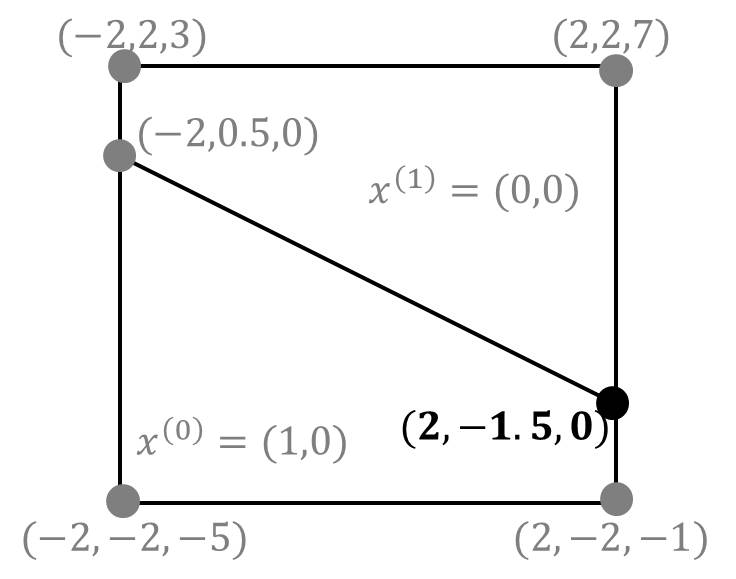}\label{fig:exe3}}
\subfigure[]{\includegraphics[width=0.45\columnwidth]{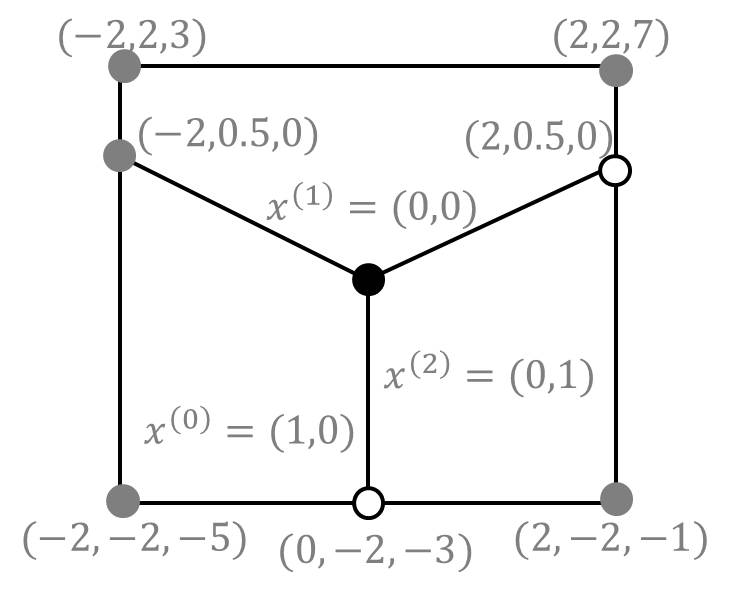}\label{fig:exe4}}
\caption{Skeletons projected onto two dimensional space. The black circle denotes the
chosen vertex in that step. The gray circle denotes that the vertex is already processed
and confirmed as a vertex of the final skeleton. The empty circle denotes a vertex not
processed yet.}
\label{fig:exe}
\end{figure}

\paragraph{Example of execution}
We explain the running process of \algname~with a toy example. Let us
consider an energy function $f(x_1,x_2) = x_1+x_2$, and two constraints $h_1$ and $h_2$
defined as follows.
\begin{align}
	h_1(x_1,x_2) &= x_1-x_2, \\
	h_2(x_1,x_2) &= 2|x_1-x_2|.
\end{align}
Here, we set $M=2$. Initially, the algorithm computes a minimizer $x^{(0)} = (1,0)$
for $\lambda^{(0)}=(-2,-2)$. Then the initial $\cV$ becomes 
$\{(-2,-2,-5), (-2,2,3),$ $(2,-2,-1), (2,2,7)\}$, which is shown in
\figurename~\ref{fig:exe1}. At this point, $X=\{x^{(0)}\}$.
Let $(-2,2,3)\in \cV$ be chosen in the next iteration, and for that vertex, the new
minimizer $x^{(1)}\in\chi_g$ is found. This updates both $X=\{x^{(0)},x^{(1)}\}$
and the skeleton as shown in \figurename~\ref{fig:exe2}. In the following iterations,
$(2,2,7),(2,-2,-1)$ and $(-2,0.5,0)$ are chosen, but for those vertices, there
is no new minimizer; that is, for those vertices, a minimizer is either
$x^{(1)}=(1,0)$ or $x^{(2)}=(0,0)$. The skeleton at this point is shown in
\figurename~\ref{fig:exe3}. Next, $(2,-1.5,0)$ is chosen, and the new minimizer
$x^{(2)}=(0,1)$ is computed so that $X$ is updated by
$\{x^{(0)},x^{(1)},x^{(2)}\}$. This changes the skeleton as in
\figurename~\ref{fig:exe4}.

\subsection{Correctness of the algorithm}

In what follows, we analyze the correctness and query complexity of \algname. All proofs
are provided in Section \ref{sec:proof}.

\blem\label{lem:skeleton_correct}
	At the end of each iteration of \algname, $\cG = \cG_{g_X}$.
\elem

Lemma \ref{lem:skeleton_correct} states that when \algname~terminates, $\cG$ is the
skeleton of an induced dual $g_X$ where $X$ is the output of the algorithm. It
remains to show that the computed $X$ is indeed the characteristic set $\chi_g$.

\bthm \label{thm:x_correct}
	When \algname~terminates with $X$, $X=\chi_g$.
\ethm
From Lemma \ref{lem:skeleton_correct} and Theorem \ref{thm:x_correct}, the following
holds.

\bcor \label{cor:skeleton_correct}
When \algname~terminates, $\cG = \cG_{\chi_g}$.
\ecor
Now we analyze the query complexity. At each iteration, the algorithm uses
exactly one oracle call.
Then, either one new $v\in \cV_g$ is identified if Line $5$ of Algorithm
\ref{alg:alg_all} is not satisfied, or one new $x\in\chi_g$ is obtained if Line
$5$ is satisfied.
Using these facts, we prove the following theorem.

\bthm \label{thm:time}
	The number of oracle calls in \algname~is $|\cV_g|+|\chi_g|$.
\ethm
\cc{Recall that each $x\in \chi_g$ corresponds to a facet of the $(m+1)$-dimensional
convex polytope of $g$. Since each vertex is determined as the intersection of
at least $(m+1)$ facets, at the end of our algorithm, $|\cV_g|$ is bounded by
$O(|\chi_g|^{m+1})$.}
Thus, the query complexity becomes $O\left(poly(|\chi_g|)\right)$.

\section{Algorithms for a specific constraint instance} \label{sec:alg_one}

In this section, we propose two \cc{variants of {\sc DualSearch}} to compute an
approximate solution for a specific constraint instance. The first one is
\cc{called} \algoneA, and the second one is \algoneB~which combines \algoneA~and
\algname. While \algname~essentially does not need prior knowledge, \algoneA~and
\algoneB~explicitly use a given prior knowledge $\hb$ for \cc{more} efficient 
computation.

\subsection{\algoneA}

Given $\hb\in\realset^m$, this algorithm finds the maximum of the dual $g$, which
provides a lower bound of \eqref{eq:our_problem}. If a corresponding minimizer
of \eqref{eq:our_problem} is in $\chi_g$, this algorithm finds that minimizer
efficiently.
Even though the corresponding minimizer is not in $\chi_g$, the algorithm finds
a lower bound of the minimum, which is a good approximate solution \cc{as shown in
Section \ref{sssec:exp}.}

The main difference of \algoneA~from \algname~is the vertex set appended to
$\cV$ in Line $7$ \cc{of Algorithm~\ref{alg:alg_all}.} At each iteration, \algoneA~calls the oracle
for the maximum of the current induced dual. While \algname~appends all vertices
in $\cV^+$, \algoneA~only appends one vertex $v\in\cV^+$ where $z_v \geq
z_u$ for all $u\in\cV^+$. Then $z_v$ becomes the maximum of the induced dual
for the next iteration. Since the (induced) dual is concave, such a local search
on $S$ enables us to eventually find the maximum of the dual. The following is
the modification of \algname~to obtain \algoneA.

\benum[\quad $\bullet$] \setlength\itemsep{-0.1\parsep}
	\item The initial vertex set is changed to $\cV' = \{v\}$ where
	$z_v \geq z_u$ for all $u\in \cV$ where $\cV$ is the ordinary initial skeleton vertex.
	\item Line \ref{line:add_v} of Algorithm \ref{alg:alg_all} is changed to ``append to
	$\cV$ the one vertex $v\in\cV^+$ such that $z_v\geq z_u$ for all $u\in\cV^+$''.
\eenum
Then, the following Lemma holds, and the proof is provided in Appendix.
\blem \label{lem:grad} When \algoneA~terminates, for the last $v^*$ for which the oracle
is called, $z_{v^*} = g(\lambda_{v^*}) = \max_\lambda g(\lambda)$.
\elem

Note that \algoneA~uses far fewer oracle calls than \algname, which
leads to fast computation of the maximum value of $g$ and the corresponding primal
solution. \cc{The cutting plane method~\cite{Guignard03} can do the same
computation as \algoneA, and \algoneA~can be understood as one implementation of
the cutting plane method.} While the cutting plane method computes the maximum of
the dual by linear programming with computed hyperplanes at each time,
\algoneA~computes it by keeping and updating the skeleton of the dual.

\cc{Now, we suggest a way for \algoneA~to deal with inequality constraints by
inserting a slack variable.} For a given problem with inequality constraints, we
first transform the problem to one with equality constraints, and apply the
algorithm to the transformed problem. Let us consider the following problem.
\beq \label{eq:ineq} 
	\min_x \left\{ f(x) : \bb-k\leq H(x) \leq \bb \right\}, 
\eeq 
where $k\in \realset^m$, and \cc{the inequality is the coordinatewise
inequality.}  The inequality gap contains our prior knowledge,
i.e.~$\hb_i\in[\bb_i-k_i,\bb_i]$. First we transform the problem to a problem
with equality constraints using a slack variable $y\in \realset^m$ as follows.
\beq \label{eq:ineq_trans} 
	\min_{x,y} \left\{ \hf(x,y) : H(x)+y = \bb \right\}, 
\eeq	
where $y\in \prod_{i=1}^m [0,k_i]$, and $\hf(x,y) = f(x)$. Let us consider the following
Lagrangian.
\beq 
	\hat{L}(x,y,\lambda) = \hf(x,y) + \lambda^T (H(x) + y - \bb).
\eeq 
For a minimizer $(x^*,y^*)$ of $\hat{L}$ for a fixed $\lambda$, it always holds that
$y^*_i = 0$ for $\lambda_i > 0$, $y^*_i = k_i$ for $\lambda_i < 0$, and $y^*$ can be any
number in $[0,k_i]$ for $\lambda_i=0$. Hence, $y^*$ only depends on $\lambda$. Then, the
dual $\hg(\lambda)$ of $\hf(x,y)$ becomes
\beq 
	\hat{g}(\lambda) = \min_{x} \left\{ f(x) + \lambda^T (H(x) + y^* - \bb) \right\}.
\eeq 
Note that $\max_\lambda \hat{g}(\lambda)$ is a lower bound of \eqref{eq:ineq}. Since
$y^*$ is determined only by $\lambda$, $\hat{g}(\lambda)$ can be computed by the same
oracle for $g(\lambda)$. Now, we obtain the following lemma.
\blem 
	Let $(x^*,y^*)$ be such that $\hat{L}(x^*,y^*,\lambda^*)=\hg(\lambda^*)$ for some
	$\lambda^*\in S$, and $b^* = H(x^*)+y^*$.
	Then $f(x^*) = \min_x \left\{ f(x) : b^*-k\leq H(x) \leq b^* \right\}$.
\elem 
\bprf 
Assume $(\hx,\hy)$ satisfying $H(\hx)+\hy = b^*$. It implies that $\lambda^T
(H(\hx)+\hy-b^*) = \lambda^T (H(x^*) + y^* - b^*)$ for any $\lambda\in \realset^m$. 
Then, $\hg(\lambda^*)=\hat{L}(x^*,y^*,\lambda^*)\leq L(\hx,\hy,\lambda^*)$. Finally,
$\hf(x^*,y^*)\leq \hf(\hx,\hy)$, and by the definition of $\hf$, $f(x^*)\leq f(\hx)$ holds.
\eprf
\cc{Hence, we can solve \eqref{eq:ineq} by the same manner as in the equality case.}
Inequality constraints make \algoneA~more widely applicable because we may not
know the exact statistics of a desired solution in practice.

\subsection{\algoneB}

\algname~is a very effective algorithm because it finds minimizers for all
$\lambda\in S$.
But in general we do not know where good solutions are found, and thus we should
use a large search region $S$, which leads to slow running time. On the other
hand, while \algoneA~efficiently finds the maximum of the dual for a specific
$\hb$, it \cc{may be} difficut to determine $\hb$ for equality constraints in
practice. Even if we use inequality constraints to deal with the uncertainty, as
inequality gap gets larger, the accuracy of \algoneA~gets lower. To overcome
these drawbacks, we propose a hybrid algorithm, called \algoneB, to combine
advantages of \algoneA~and \algname, which runs as follows.

\benum[Step $1$] \setlength\itemsep{-0.1\parsep}
	\item Let our prior knowledge $\hb$ be given, and $S\subset\realset^m$ be a large search
	region.
	\cc{
	\item Run \algoneA~on $S$ with inequality constraint $\hb-k^-\leq H(x)\leq \hb+k^+$ for 
	moderately small $k^-,k^+ > 0$. 
	Let $b^*$ be the constraint instance for which the dual maximum is computed.
	\item  Run \algoneA~again on $S$ with equality constraint $H(x)=b^*$. Then, we obtain 
	$\lambda^*$ at which \algoneA~computes the maximum of the dual.
%
}
	\item Run \algname~for a small search region $\prod_{i=1}^m
	[\lambda^*_i-\alpha_i,\lambda^*_i+\alpha_i]$ where $\alpha_i\geq 0$, and let $X^*$ be the
	output of \algname.
	\item Output a solution among $X^*$ that minimizes the soft-constrained objective.
\eenum
Note that in \algoneB, we can also use the cutting plane method instead of
\algoneA. In general, any convex search region is adoptable in Step $4$, but we observed
from extensive experiments that small constants $\alpha_i$ are enough to obtain a good
solution. We will show in Section~\ref{sec:app} that \algoneB~computes better solutions
than \algoneA~and runs quite fast.


\section{Applications}\label{sec:app}

\subsection{Labelling problems in computer vision}\label{ssec:imgseg}

In computer vision, a number of problems can be reduced to labelling problems, including
image segmentation, $3$D-reconstruction, and stereo. Our constrained energy minimization
algorithms can be applied to those problems, for instance, when we may have knowledge on
the volume of a reconstructed object for $3$D-reconstruction or on the number of pixels
belonging to an object for image segmentation. In this section, we show how our algorithms
are applied to the foreground-background (fg-bg) image segmentation problem.

The fg-bg image segmentation problem is to divide a given image to foreground (object) and
 background. This can be done by labelling all pixels such that $1$ is assigned to
foreground pixels and $0$ is assigned to background pixels. For this problem, one popular
approach is to consider an image as a grid graph in which each node has four neighbours,
and minimize an energy function $f$ of the form \eqref{eq:pairwise_energy_function},
which is submodular.
The unary terms of the function encode how likely each pixel belongs to the foreground or
background, while the pairwise terms encode the smoothness of the boundary of the object
being segmented.
However, in general, a minimizer of \eqref{eq:pairwise_energy_function} is not the ground
truth, and it has been shown that imposing statistics on a desired solution can improve
segmentation results~\cite{KlodtC11}.


Below, we describe some linear constraints that have been successfully used in
computer vision.
\begin{enumerate}[$\bullet$] 
	\item {\bf Size:} $\sum_{i\in V} x_i=b$ where $b\in \realset$ \cite{LimJK10,wernercvpr08,woodfordiccv09}.
	\item {\bf Mean:} $\sum_{i\in V} \frac{c_ix_i}{\sum_{i\in V}x_i}=b$ where $b\in \realset^2$ and 
	$c_i = (v_i, h_i)\in\realset^2$ denotes the vertical and horizontal coordinates of a pixel $i$, 
	respectively \cite{KlodtC11}.
	\item {\bf Cov.:} $\sum_{i\in V} \frac{(v_i-\mu_v)(h_i-\mu_h)x_i}{\sum_{i\in V}x_i}=b$ where 
	$b\in \realset$ and $(\mu_v,\mu_h)\in\realset^2$ denotes the mean center of the object \cite{KlodtC11}.
\end{enumerate}
We can define the variance constraints for the vertical and horizontal
coordinates in a similar way to the covariance constraint.

\begin{figure}
\centering
\includegraphics[width=0.95\columnwidth]{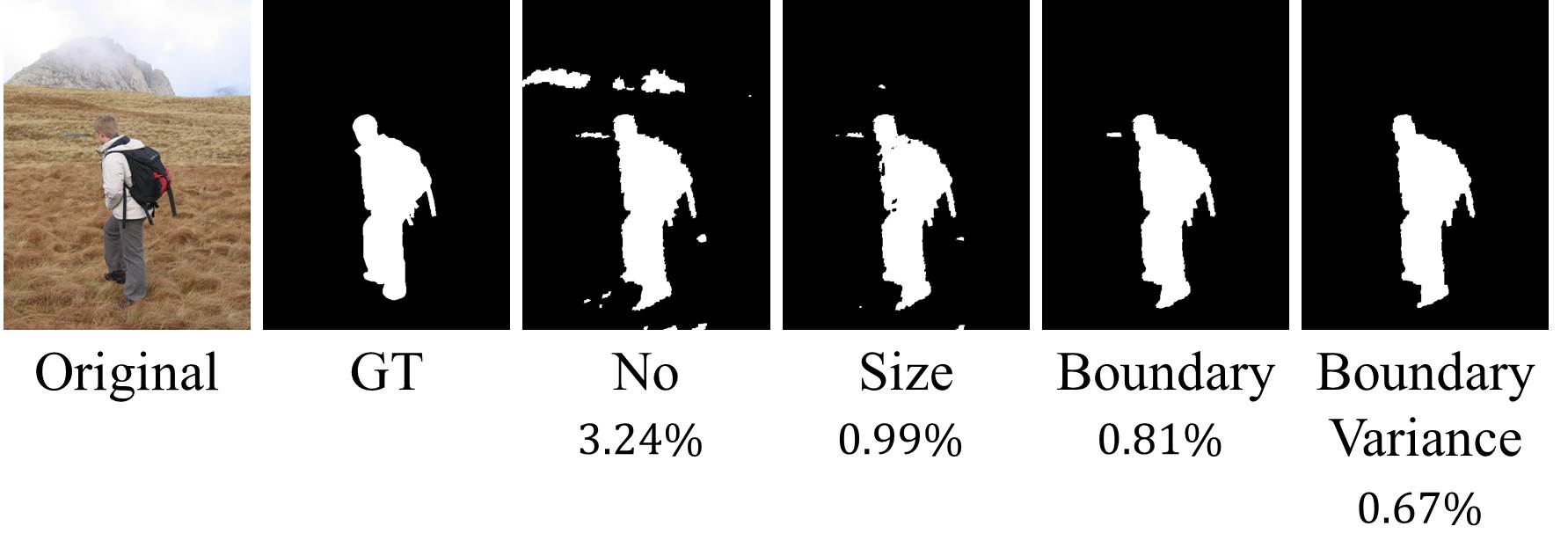}
\caption{The segmentation labelled by No is obtained by
minimizing the energy function with no constraint. The last three segmentations were
obtained by \algoneA~with the specified constraints whose instances are set by the ground
truth statistics. For each segmentation, the pixel-by-pixel error with respect to the
ground truth is reported.}
\label{fig:segmentation}
\end{figure}

In many scenarios, researchers are interested in ensuring that the boundary of
the object in the segmentation has a particular length. This length can be
measured by counting the number of pairs of adjacent variables having different
labels and described by $\sum_{(i,j)\in E}|x_i-x_j| = b$ where $b\in \realset$.
For this {\bf boundary constraint}, the search region $S$ may be restricted to a
subregion of $\realset\times [K,\infty]$ where $K\leq 0$ is the smallest real
number ensuring $L(x,\lambda)$ submodular for all $\lambda\in S$.
\figurename~\ref{fig:segmentation} shows improvement of segmentation results by
imposing the above constraints.



\subsubsection{Query complexity of \algname}
Recall that the query complexity of \algname~is polynomial in $|\chi_g|$. \cc{Note that
$|\chi_g|$ is upper bounded by the number $C$ of all possible constraint instances.
For all the constraints above, we can show $C=O(poly(n))$.} 
For example, for the size constraint, $C=n$, and for the boundary length
constraint, $C\leq 2n$ because $G$ is a grid graph.
\cc{Let us consider the mean constraint, and let $\hb\in\realset^2$ be obtained from our
 prior knowledge.}
Then, the Lagrangian is as follows:
\begin{equation}
	L(x,\lambda) = f(x) + \lambda^T \left(\sum_i (c_i-\hb)x_i \right),
\end{equation}
where $c_i\in\intset^2$ is bounded by the size of row and column of the image. Hence, the
numbers of possible values of $\sum_i c_ix_i$ and $\hb\sum_i x_i$ are $O(n^2)$ and $O(n)$,
respectively, which leads to $C=O(n^3)$.
\cc{By a similar analysis, we can show $C=O(n^3)$ for the covariance and variance constraints. 
If we consider multiple
constraints simultaneously, $C$ is bounded by multiplication of the upper bound
of each constraint.
Hence, $|\chi_g| = O(poly(n))$ for any combination of the constraints above.}

\subsubsection{Experiments} \label{sssec:exp}

\cc{First we did experiments for the size and boundary constraints, and used the following Lagrangian.}
\beq \label{eq:imgseg_lagrangian} L(x,\lambda) = f(x) + \\ \lambda_1 \sum_{i\in V}  x_i +
\lambda_2 \sum_{(i,j)\in E} |x_i-x_j|.
\eeq


\begin{table}[t]
\centering
\caption{Results of \algname~on $12$ images from \cite{RhemannRRS08} each of which
has the size $120\times 120$.}
\begin{tabular}{lccc|lccc} \hline
	&  $|\cV_g|$ & $|\chi_g|$ & Time & & $|\cV_g|$ & $|\chi_g|$ & Time \\ [0.2ex] \hline\hline \\ [-2.2ex]
	IM1 & 273K & 286K & 25m & IM5 & 306K & 325K & 31m \\
	IM2 & 168K & 170K & 16m & IM6 & 238K & 252K & 27m\\
	IM3 & 105K & 107K & 14m & IM7 & 248K & 248K & 23m\\
	IM4 & 114K & 127K & 15m & IM8 & 300K & 308K & 31m\\
	\hline
\end{tabular}
\label{tab:result_summary}
\end{table}

\begin{figure}
\centering
	\includegraphics[width=0.6\columnwidth]{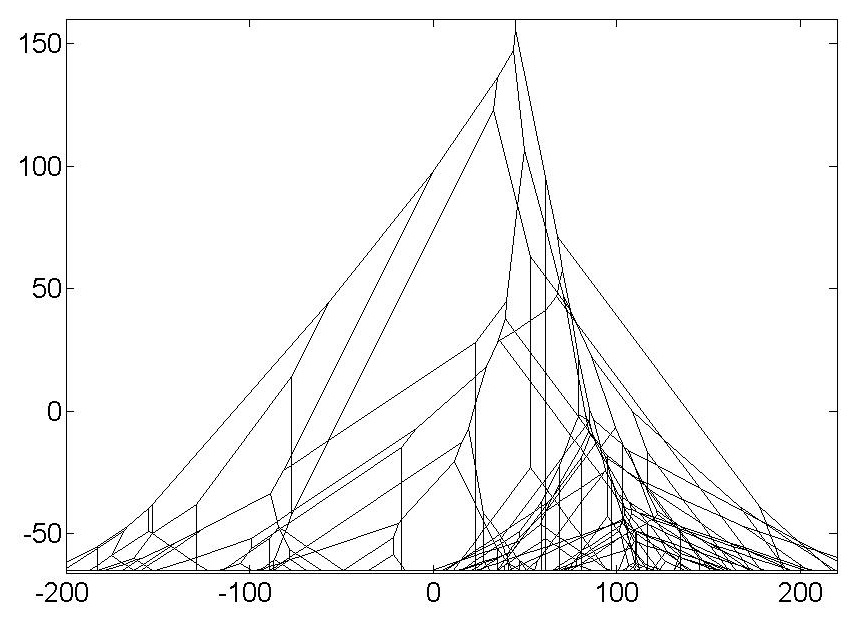}
	\caption{Example of a skeleton projected onto two dimensional space, computed with \eqref{eq:imgseg_lagrangian}.}
	\label{fig:skeleton_ex}
\end{figure}

\tablename~\ref{tab:result_summary} reports the summary of results of \algnamec for $12$
images with size $120\times120$. \algname~produces minimizers for a very large number of
constraint instances.
\cc{One implication is that for any given constraint instance, \algoneAc and \algoneBc 
can compute a minimizer with very close constraint instance to the original one.} 
\figurename \ref{fig:skeleton_ex} shows an example of a skeleton projected onto two dimensional $\lambda$ space that is computed with \eqref{eq:imgseg_lagrangian} for a $12\times 12$ image.

\begin{figure}
\centering
\includegraphics[width=0.95\columnwidth]{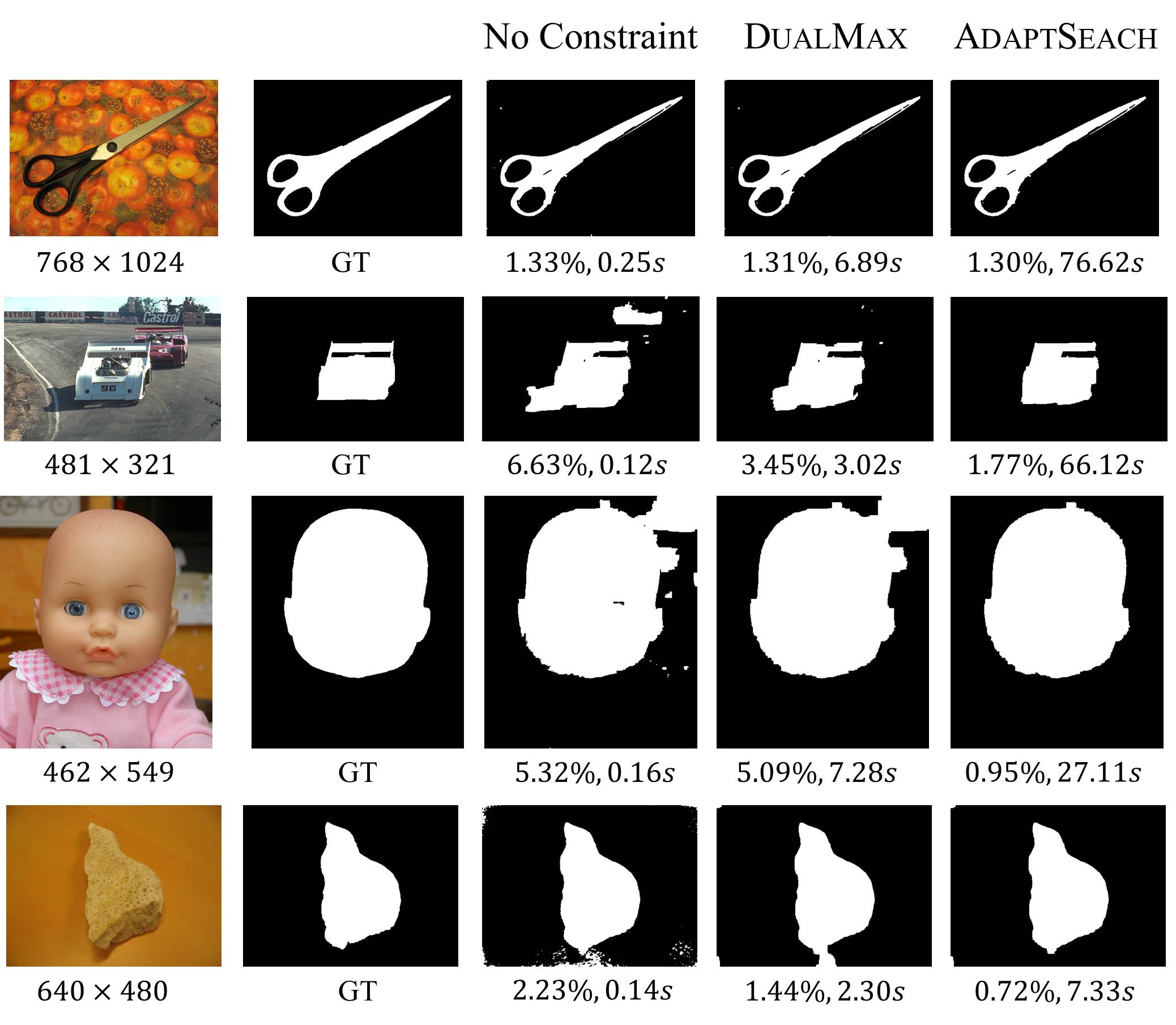}
\caption{Segmentation results by minimizing
\eqref{eq:pairwise_energy_function}, \algoneAc and \algoneBc each of which is labelled by
the pixel-by-pixel error with respect to the ground truth and its running time. The used
images are from \cite{RotherKB04}.}
\label{fig:method_for_one}
\end{figure}

\figurename~\ref{fig:method_for_one} shows experimental results of \algoneA~and
\algoneB.
For \algoneB, we used \cc{a soft constraint with} a square penalty function for
the size and the boundary length constraint, that is, $\eta_1 (\sum_i x_i -
\hb_1)^2$ and $\eta_2 ( \sum_{(i,j)\in E } |x_i-x_j| - \hb_2)^2$. We chose
$\eta_1=1$ and $\eta_2 = 100$ with which segmentation results generally show
less error. Also for the first running of \algoneA, we used the inequality gap
$k^+,k^-$ of $\pm 10\%$ of $\hb$, and $\hb_1,\hb_2$ were obtained from the ground
truth.
The small search region to apply \algname~is used with $\alpha_1=\alpha_2=1$,
except for the first image with $\alpha_1=\alpha_2=0.3$.

\cc{We also compared our algorithms with LP~\cite{lempitskyiccv09} and QP~\cite{KlodtC11} 
relaxation based methods.} 
\tablename~\ref{tab:comparison_with_cont} shows that
\algoneA~is faster and more accurate compared with both methods. Since LP and QP
cannot handle higher order both-side constrained inequality constraints \cc{unlike 
our algorithms}, we used
linear constraints introduced previously.
Segmentation results are provided in Appendix.

 \renewcommand{\arraystretch}{1.1}
\begin{table}[t]
\caption{Comparison of \algoneA~and two continuous relaxation based methods with
inequality gap $\pm 5\%$.
The values are averaged over 6 images from \cite{RotherKB04} of size $321\times 481$.}
\centering
\begin{tabular}{lcccccc}  \hline \\ [-2.5ex]
	\multirow{2}{8mm}{Const.} & \multicolumn{2}{c}{\algoneA} & \multicolumn{2}{c}{LP} &
	\multicolumn{2}{c}{QP} \\ \cline{2-7} \\ [-2.5ex]
		  & Err. & Time & Err. & Time & Err. & Time \\ \hline\hline \\ [-2.5ex]
	Sz,Mn & 3.37 & 1.73 & 3.46 & 83.6 & 3.95 & 349 \\	
	Sz,Vr & 2.34 & 1.91 & 2.73 & 77.2 & 2.64 & 461 \\
	\hline
\end{tabular}
\begin{flushleft}
\small
Err.: pixel-by-pixel error with respect to the ground truth $(\%)$. \\
Time: in seconds. \qquad Const.: constraints. \\
Sz: size, Mn: Mean, Vr: variance.
\end{flushleft}
\label{tab:comparison_with_cont}
\end{table}
\renewcommand{\arraystretch}{1.0}

\subsection{Combinatorial optimization} \label{ssec:combinatorial_opt}

\paragraph{Submodular minimization}
Our method can also be used for constrained submodular function minimization
(SFM). SFM is known to be polynomial time solvable and a number of studies have
considered SFM under specific constraints such as vertex cover and size
constraints~\cite{IwataN09,NaganoKA11}. In contrast to previous work, we
provide a framework for dealing with multiple general constraints. Our method
can not only deal with any linear constraint, but can also handle some higher
order constraints which ensure that the dual is computable. For instance, as
shown in the previous section, any submodular constraint $h_i(x)$ can be handled
with restricted $\lambda_i\geq 0$.

\paragraph{Shortest path problem}
The restricted shortest path problem is a widely studied constrained version in
which each edge has an associated delay in addition to its length. A path is
feasible if its total delay is less than some threshold $D$~\cite{LorenzR99}.
This is a linear constraint $\sum_i d_ix_i\leq D$ where $d_i$ is the delay of
edge $i$. Another natural constraint for the shortest path problem is to drop
some $k$ nodes among a given set of $m$ nodes. For instance, we may want to
design a tour that should contain $k$ cities among $m$ cities. Indeed, this
becomes a Hamiltonian path problem when $k=m=n$. As in the project selection
problem, we may partition $n$ cities to $r$ groups, and try to visit $k_i$
number of cities from each group where $1\leq i\leq r$. Note that all
constraints above are linear so that our method can be applied.

\paragraph{Project selection problem}
Given a set $P$ of projects, a profit function $q:P\rightarrow \realset$, and a
prerequisite relation $R\subseteq P\times P$, this problem is to find projects
maximizing the total profit while satisfying a prerequisite relation. This is
also known as the maximal closure problem and can be solved in polynomial time
by transforming it to a st-mincut problem~\cite{Picard76}. In practice, $P$ may
be represented by sets $P_1,\ldots,P_m$ that may overlap, and we may want to
select $k_i$ projects from $P_i$ for $1\leq i\leq m$. This can be formulated
using linear constraints $k_i = u_i^Tx$ where $u_i$ is an indicator which
projects belong to $P_i$. This enables the use of our method to solve the
constrained project selection problem.

\section{Conclusions}
This paper proposes novel algorithms to deal with the multiple constrained MAP inference
problem. Our algorithm \algoneB~is able to generate high-quality candidate
solutions in a short time (see \figurename~\ref{fig:method_for_one}) and enables handling
of problems with very high order potential functions. We believe it would have a
significant impact on the solution of many labelling problems encountered in computer
vision and machine learning. As future work, we intend to analyze the use of our
algorithms for enforcing statistics in problems encountered in various domains of machine
learning.



\bibliographystyle{abbrv}
\bibliography{ref}

\newpage

\setcounter{section}{0}

\renewcommand{\thesection}{\Alph{section}}

\noindent
{\bf\LARGE Appendix}

\section{Proofs} \label{sec:proof}

In this section, we provide the proofs omitted in the main body. We use notations $\cG(t),$ $\cV^+(t),$ $\cV^-(t)$ and $X(t)$ to indicate $\cV^+,\cV^-,\cG$ and $X$ at the end of the $t$-th iteration in lgorithm~\ref{alg:alg_all}, respectively. Also we denote a vertex chosen in Line $3$ at the $t$-th iteration by $v$, and $\Gamma_X(S)$ by $\Gamma_X$ without $S$. Note that initially $\cG(0) = \cG_{g_{X}}$ by the definition of $InitSkeleton()$.

\subsection{Proof of Lemma \ref{lem:skeleton_correct}}


Lemma \ref{lem:skeleton_correct} is proved by the following 
Lemma \ref{lem:skeleton_v_correct} and Lemma \ref{lem:skeleton_e_correct}.

\blem \label{lem:skeleton_v_correct}
	Assume that $\cG(t-1) = \cG_{g_{X(t-1)}}$. Then $\cV(t)=\cV_{g_{X(t)}}$.
\elem
\bprf
	$(\Longrightarrow)$
	Let $u\in \cV(t)$. First assume that $u$ is also in $\cV(t-1)$. 
	Suppose that $u \notin \cV_{g_{X(t)}}$. 
	There is $Q\subseteq \Gamma_{X(t)}$ and $Q \cap \{u\} = \emptyset$ so that $u$ is 
	a proper convex combination of $Q$. Note that $\Gamma_{X(t)}\subseteq \Gamma_{X(t-1)}$. 
	Thus, $u$ is a proper convex combination of $Q$ over $\Gamma_{X(t-1)}$. 
	It is a contradiction to $u\in \cV(t-1)$.
	
	Assume that $u\notin \cV(t-1)$. Then, $u\in \cV^+(t)$, implying that 
	$u\in e(u_1,u_2)\in \cE(t-1)$ and $u\in P_{x_v}$. 
	Suppose that there is $Q\subseteq \Gamma_{X(t)}$ and $Q \cap \{u\} = \emptyset$ 
	so that $u$ is a proper convex combination of $Q$. Since $Q\nsubseteq e(u_1,u_2)$ and 
	$\Gamma_{X(t)}\subseteq \Gamma_{X(t-1)}$, it is a contradiction to the definition of 
	$\cE(t-1)$.
	
	

	$(\Longleftarrow)$
	Let $u\in \cV_{g_{X(t)}}$. Suppose that $u\notin \cV(t)$ but $u\in \cV(t-1)$, 
	which means that $u \in \cV^-$. 
	Then, $z_u > g_{X(t)}(\lambda_u)$, a contradiction to $u\in \cV_{g_{X(t)}}$.	
	Suppose that $u\notin \cV(t)$ nor $u\notin \cV(t-1)$. 
	Then, there is $Q\subset \Gamma_{X(t-1)}$ and $Q\cap \{u\} = \emptyset$ 
	so that $u$ is a proper convex combination of $Q$.	
	Note that $Q\nsubseteq P_{x_v}$ because $u\in\cV_{g_{X(t)}}$. 
	Since $z_u \leq P_{x_v}(\lambda_u)$, 
	at least one of $Q$ is strictly below $P_{x_v}$, and 
	let $Q^-$ be the set of such elements of $Q$.	
	Since $u\in \cV_{g_{X(t)}}$, 
	at least one of $Q$ is strictly above $P_{x_v}$, and 
	let $Q^+$ be the set of such elements of $Q$.
	Let $P$ be the set of intersections of $e(q^-,q^+)$ and 
	$P_{x_v}$ where $q^-\in Q^-$ and $q^+\in Q^+$. 
	Suppose that $u$ is strictly below $P_{x_v}$, then $u$ is a proper convex 
	combination of $P$ and $Q^-\subset \Gamma_{X(t)}$, implying a contradiction. So $u\in P_{x_v}$.
	Suppose that $|P|>1$, then $u$ is a proper convex combination of $P\subset \Gamma_{X(t)}$, 
	which is a contradiction.
	Thus, $|P| = 1$ and $u\in P_{x_v}$. Then since $u$ is on some edge $e(w_1,w_2)\in\cE(t-1)$ 
	and $u\in P_{x_v}$, $u\in \cV^+(t)$ by the algorithm so that $u$ is present in $\cV(t)$, which 
	is a contradiction. 	
\eprf

\blem \label{lem:skeleton_e_correct}
	Assume that $\cG(t-1) = \cG_{g_{X(t-1)}}$. Then $\cE(t)=\cE_{g_{X(t)}}$.
\elem
\bprf
	$(\Longrightarrow)$
	Let $e(u,w) \in \cE(t)$.
	Assume that $e(u,w)$ is added by $\cE^+$ so that 
	$w$ is the intersection of $e(u,u')\in \cE(t-1)$ and $P_{x_v}$. 
	Suppose that there is $Q\subset \Gamma_X(t)$, and $Q\nsubseteq e(u,w)$ so that 
	for some $p\in e(u,w)$, $p$ is a proper convex combination of $Q$. 
	Since $\Gamma_{X(t)}\subseteq \Gamma_{X(t-1)}$, $Q\subseteq \Gamma_{X(t-1)}$. 
	Also since $Q\subset \Gamma_{X(t)}$ and $Q\nsubseteq e(u,w)$, $Q\nsubseteq e(u,u')$. 
	It is a contradiction to $p\in e(u,u')\in\cE(t-1)$.
	

	Assume that $e(u,w) \in \cE(t-1)$. 
	Since $\Gamma_{X(t)}\subseteq \Gamma_{X(t-1)}$, no $p\in e(u,w)$ is a proper convex 
	combination of $Q\in \Gamma_{X(t)}$ and $Q\nsubseteq e(u,w)$. 
	Thus, $e(u,w) \in \cE_{g_{X(t)}}$ due to $u,w\in \cV(t) = \cV_{g_{X(t)}}$ by 
	Lemma \ref{lem:skeleton_v_correct}.

	
	Assume that $e(u,w)$ is added by $ConvEdge(\cV^+(t))$.
	Suppose that there is $Q\subset \Gamma_{X(t)}$ and $Q\nsubseteq e(u,w)$ so that 
	for some $p\in e(u,w)$, $p$ is a proper convex combination of $Q$. 
	Since $p\in e(u,w) \in ConvEdge(\cV^+(t))\subset P_{x_v}$, 
	$Q\subset \Gamma_{X(t)} \cap P_{x_v}$. 
	Since $e(u,w)$ is an edge of the convex hull of $\cV^+(t)$, any $p\in e(u,w)$ cannot
	be a proper convex combination of $Q$, which is a contradiction.
	
	

	$(\Longleftarrow)$ 
	Let $e(u,w)\in \cE_{g_{X(t)}}$. Suppose that $e(u,w) \notin \cE(t)$. 
	If $u\in \cV^-$ or $w\in \cV^-$, it is a contradiction to 
	$e(u,w)\in \cE_{g_{X(t)}} \subset 2^{\Gamma_{X(t)}}$. 
	If both $u,w\in \cV^+(t)$, by the definition of $ConvEdge(\cV^+)$, and the fact 
	that $e(u,w)\notin \cE(t)$, $e(u,w)\notin \cE_{g_{X(t)}}$, which is a contradiction. 
	Therefore one of $u,w$ belongs to $\cV(t-1)\cap \cV(t)$, and 
	the other belongs to $\cV^+(t)$. 
	Without loss of generality, let $u\in \cV(t-1)\cap \cV(t)$ and 
	$w\in \cV^+(t)$, then $u$ must be strictly below $P_{x_v}$.	
	Then, $e(u,w)\notin \cE(t-1)$. 
	There is $Q\in\Gamma_{X(t-1)}$ so that $Q\nsubseteq e(u,w)$ and 
	for some $p\in e(u,w)$, $p$ is a proper convex combination of $Q$. 
	If all $q\in Q$ are strictly above $P_{x_v}$, $p$ is also strictly above $P_{x_v}$. 
	If for all $q\in Q$, $z_q \leq P_{x_v}(\lambda_q)$, $Q\subset \Gamma_{X(t)}$, implying 
	a contradiction to $e(u,w)\in \cE_{g_{X(t)}}$.
	Thus, at least one of $Q$ is strictly below $P_{x_v}$, and let $Q^-$ be the set 
	of such elements of $Q$.
	Also at least one of $Q$ is strictly above $P_{x_v}$, and let $Q^+$ be the set of such 
	elements of $Q$. 
	Let $P$ be the set of intersections of $e(q^-,q^+)$ and $P_{x_v}$ 
	where $q^-\in Q^-$ and $q^+\in Q^+$. 
	If $|P|>1$ or $p\notin P$, $p$ is a proper convex combination of $Q^-$ and $P$. 
	Thus, $|Q^-|=|Q^+|=1$ and $p\in P_{x_v}$. This holds for all $p\in e(u,w)$, 
	which means that $e(u,w) \in P_{x_v}$.  This is a contradiction to the fact 
	that $u$ is strictly below $P_{x_v}$.
\eprf

\subsection{Proof of Theorem \ref{thm:x_correct}}

\bprf
	Let $\cG=(\cV,\cE)$ be the skeleton at the end of \algname.
	It holds that $X\subseteq \chi_g$ for each iteration by the algorithm. 
	Suppose that there is $x^* \in \chi_g\backslash X$ when the algorithm terminates.
	Then, there is $\lambda^*\in S$ such that $P_{x^*}(\lambda^*) < P_{x}(\lambda^*)$ 
	for every $x \in X$. 
	Also there is $\hx\in X$ such that $P_{\hx}(\lambda^*) = g_X(\lambda^*)$. 
	Then, $(\lambda^*,P_{\hx}(\lambda^*))$ can be represented as 
	a convex combination of $\cV'\subset \cV$ 
	such that $P_{\hx}(\lambda_v)=g_X(\lambda_v)$ for all $v\in \cV'$. 
	By a property of the algorithm, 
	$P_{x^*}(\lambda_v) \geq P_{\hx}(\lambda_v) = g_X(\lambda_v) = g(\lambda_v)$ 
	for each $v\in \cV'$. 
	Since $\lambda^*$ is a convex combination of $\{\lambda_v : v\in \cV'\}$, 
	we have $P_{x^*}(\lambda^*) \geq P_{\hx}(\lambda^*)$. This implies a contradiction 
	to $P_{x^*}(\lambda^*) < P_{\hx}(\lambda^*)$.	
\eprf

\subsection{Proof of Theorem \ref{thm:time}}
\bprf
	For each iteration, there is exactly one oracle call. Let $C$ be a set of 
	confirmed vertices $u$, that is, $z_u = g(\lambda_u)$. 
	Note that at each iteration, either $|C|$ or $|X|$ increases by one, 
	depending on whether $P_{x_v}(\lambda_v) < z_v$, and confirmed 
	vertices are never removed from $\cV$.
	When the algorithm terminates, $|X|=|\chi_g|$ by Theorem \ref{thm:x_correct} and 
	$|C| = |\cV_g|$ by Corollary \ref{cor:skeleton_correct}. Thus, the algorithm 
	uses $|\cV_g|+|\chi_g|$ number of oracle calls.
\eprf

\subsection{Proof of Lemma \ref{lem:grad}}


\bprf
	First we prove that for every iteration, a chosen $v$ satisfies that 
	$z_v = \max_\lambda g_{X}(\lambda)$. It initially holds by the definition of $v$. 
	Assume that at the $(t-1)$-th iteration, the statement holds. 
	Let $v$ be chosen in the $t$-th iteration. If $P_{x_v}(\lambda_v) \geq z_v$, 
	there is no change on $X$ and $\cG$ so that the statement holds, and the 
	algorithm terminates.
	When $P_{x_v}(\lambda_v) < z_v$, let $\hv\in\cV^+$ be such that $z_{\hv} \geq z_u$ 
	for all $u\in \cV^+$. 
	Suppose that there is $\bv\in \cV(t)$ such that $z_{\bv} > z_{\hv}$ and $z_{\bv}$ 
	is the maximum value of $g_{X(t)}$. 
	Note that $z_v$ is the maximum value of $g_{X(t-1)}$ by the assumption.	
	Suppose $z_{\bv} = z_{v}$, and let $P\subseteq\cV(t-1)$ be the set such that for every
	$p\in P$, $z_p = z_v$.
	Since $\bv\in \cV(t)$, $z_{\bv} \leq P_{x_v}(\lambda_{\bv})$ and $z_v >
	P_{x_v}(\lambda_v)$.
	Then, some edge between two vertices of $P$ should intersect with $P_{x_v}$ 
	due to the concavity of $g_{X(t)}$, and let $v'$ be the intersection.
	Then, $z_v=z_{v'}=z_{\hv}$, which is a contradiction to $z_{\bv} > z_{\hv}$.

	In $\cG_{X(t-1)}$, since $\bv$ is not the maximum, and by the concavity of 
	$g_{X(t-1)}$, there is at least one edge $e(\bv,\bv')$ where $\bv'\in \cV(t-1)$ 
	such that $z_{\bv'} > z_{\bv}$. 
	In order that $z_{\bv}$ becomes the maximum of $g_{X(t)}$, $\bv'$ should not 
	belong to $\cV(t)$, implying that $z_{\bv'} > P_{x_v}(\lambda_{\bv'})$.
	Then, there is intersection $q$ of $P_{x_v}$ and $e(\bv,\bv')$, implying that
	$z_q\geq z_{\bv}$. If $z_q > z_{\bv}$, it is a contradiction to the fact that 
	$z_{\bv}$ is the maximum value of $g_{X(t)}$ due to $q\in \cV(t)$. 
	If $z_q = z_{\bv}$, it means that $z_{\bv}\in\cV^+$ and thus $z_{\bv} = z_{\hv}$, 
	which is a contradiction to $z_{\bv}>z_{\hv}$.
	 
	
	We have proved that when the algorithm terminates, 
	$z_v = \max_\lambda g_X(\lambda)\geq \max_\lambda g(\lambda)$. 
	Let the last $v$ be $v^*$. 
	In the last iteration, 
	$z_{v^*} = P_{x_{v^*}}(\lambda^*) = g(\lambda^*)\leq \max_\lambda g(\lambda)$.
\eprf


\newpage
\section{Segmentation results for Table \ref{tab:comparison_with_cont}}

\begin{center}
\includegraphics[width=0.9\columnwidth]{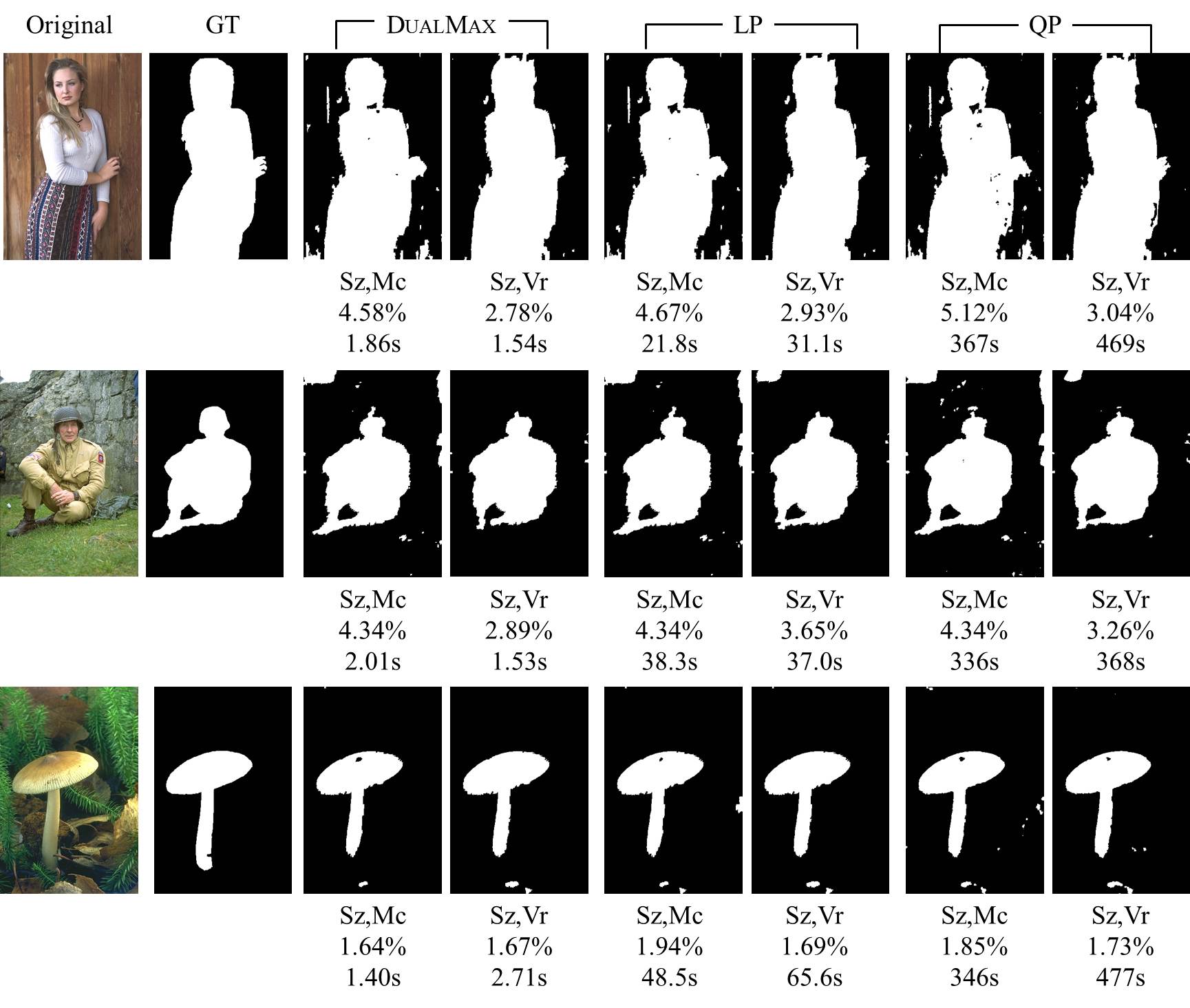}
\includegraphics[width=1\columnwidth]{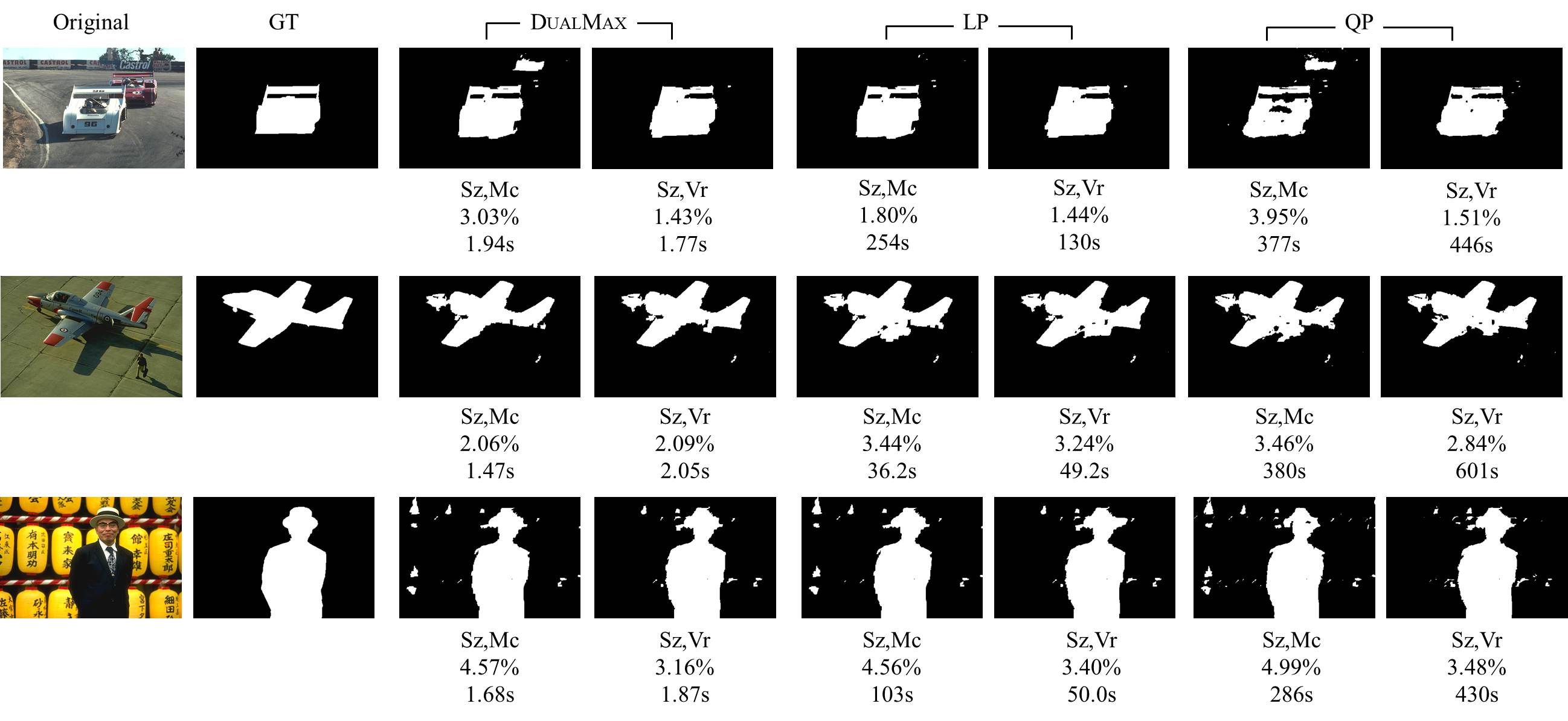}
\end{center}

%

\end{document}